\documentclass[journa]{IEEEtran}

\usepackage{graphicx}
\usepackage{amsmath,amssymb} 
\usepackage{color}
\usepackage{epstopdf}
\usepackage{subfigure}
\usepackage{placeins}
\usepackage{url}
\usepackage{tablefootnote}
\usepackage{pbox}

\usepackage[pagebackref=true,breaklinks=true,letterpaper=true,colorlinks=true,bookmarks=false]{hyperref}

\usepackage{algorithmic}

\DeclareMathAlphabet{\mathpzc}{OT1}{pzc}{m}{it} 

\newcommand{\etal}{et al.\!}
\newcommand{\eg}{e.g.\!}
\newcommand{\ie}{i.e.\!}

\ifCLASSINFOpdf

\else

\fi

\begin{document}
\title{Learning to Recognize Pedestrian Attribute}

\author{Yubin~Deng,
        Ping~Luo,
        Chen~Change~Loy,~\IEEEmembership{Member,~IEEE,}
        and~Xiaoou~Tang,~\IEEEmembership{Fellow,~IEEE}
\thanks{}%
\thanks{Y. Deng, P. Luo, C. C. Loy and X. Tang are with the Department
of Information Engineering, The Chinese University of Hong Kong, Hong Kong.
 (e-mail: danny.s.deng.ds@gmail.com;
lp011@ie.cuhk.edu.hk;
ccloy@ie.cuhk.edu.hk;
xtang@ie.cuhk.edu.hk)}
\thanks{}

}

\markboth{}%
{Shell \MakeLowercase{\textit{et al.}}: Bare Demo of IEEEtran.cls for Journals}
%

\maketitle

\begin{abstract}
Learning to recognize pedestrian attributes at far distance is a challenging problem in visual surveillance since face and body close-shots are hardly available; instead, only far-view image frames of pedestrian are given. In this study, we present an alternative approach that exploits the context of neighboring pedestrian images for improved attribute inference compared to the conventional SVM-based method. In addition, we conduct extensive experiments to evaluate the informativeness of background and foreground features for attribute recognition.
Experiments are based on our newly released pedestrian attribute dataset, which is by far the largest and most diverse of its kind.
\end{abstract}


%
\IEEEpeerreviewmaketitle

\section{Introduction}
%
%
%
%
\IEEEPARstart{L}{earning} to recognize pedestrian attributes, such as gender, age, clothing style, has received growing attention in computer vision research, due to its high application potential in areas such as video-based business intelligence~\cite{VideoBased} and visual surveillance~\cite{GongChallenge2013}. In real-world video surveillance scenarios, clear close-shots of face and body regions are seldom available. Thus, attribute recognition has to be performed at far distance using pedestrian body appearance (which can be partially occluded) in the absence of critical face/close-shot body visual information.

\begin{figure}[t]
\centering
\includegraphics[width=0.48\textwidth]{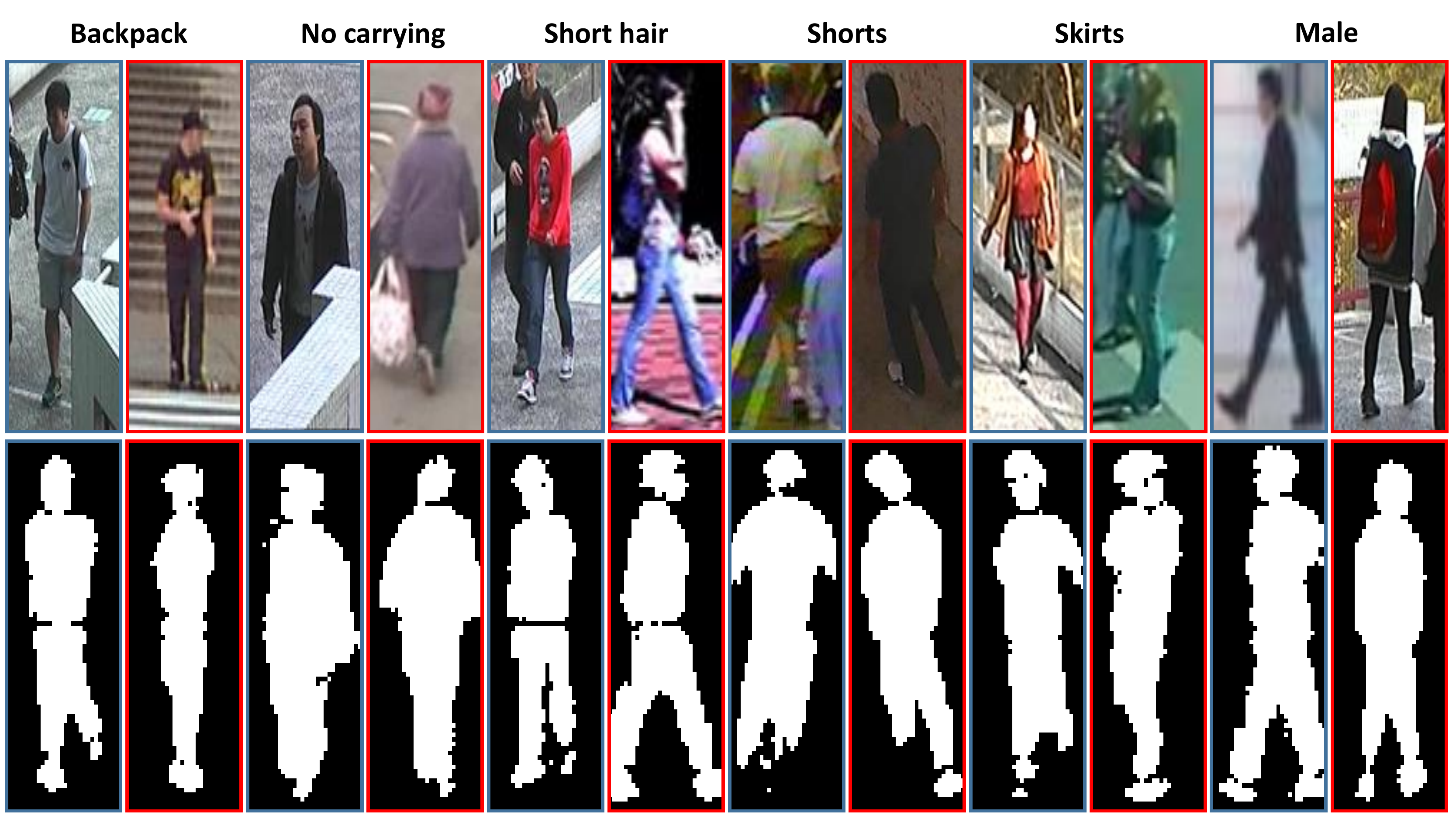}
\vskip -0.3cm
\caption{Sample images of far-view pedestrian and their corresponding binary parsing masks. Positive and negative samples are indicated by blue and red boxes, respectively.}
\vskip -0.5cm
\label{fig:figParsed}
\end{figure}

Pedestrian attribute recognition at far distance is non-trivial due to:
1) \textit{Appearance diversity} - owing to diverse appearances of pedestrian clothing and uncontrollable multi-factor variations such as illumination and camera viewing angle, there exist large intra-class variations among different images for the same attribute;
%
%
%
%
2) \textit{Appearance ambiguity} - far-view attribute recognition is a remarkably difficult task due to limited image resolution, inherent visual ambiguity, and poor quality of visual features obtained from far view field (Fig.~\ref{fig:figParsed}). 

\vspace{0.1cm}
\noindent \textbf{Related work:}
Cao~\etal~\cite{cao2008gender} are among the first to study human attribute recognition from full body images. In their study, HOG features extracted from overlapping patches are used along with Adaboost classifier for recognizing the gender attribute.
%
%
Bourdev~\etal~propose the use of poselets~\cite{bourdev2011describing} to attribute recognition. In particular, HOG features, color histogram, and skin-specific
features are extracted on local poses for poselet-level attribute classification.
%
Zhu~\etal~\cite{Zhu} extract dense color, LBP, and HOG features to train Adaboost and weighted $k$NN classifiers for attributes classification.
Although these approaches have all tried to train a robust attribute detection model, they either relied on a small-size dataset or selected not enough attributes for analysis. In view of the growing research interest in the field of human re-identification~\cite{GongChallenge2013,zheng2013reidentification,liu2014semi}, which aims at detecting the same person across spatial and temporal distance, the role of pedestrian attributes has become vital, as mid-level features are shown to be exceptional for aiding the human re-identification task~\cite{liu2012person}. 
In particular, Layne~\etal~\cite{layne2012towards,Layne} propose intersection kernel SVM with a mixture of colour (RGB, HSV and YCbCr) and texture histograms (8 Gabor filters and 13 Schmid filters) for learning a selection of pedestrian attributes as a form of mid-level features to describe people. The use of attributes has shown remarkable re-identification performance compared to employing low-level features alone, but the attribute recognition performance in~\cite{layne2012towards,Layne} has yet to be improved.

%

\vspace{0.1cm}
\noindent \textbf{Contributions:}
As discussed above, most existing pedestrian attribute studies focus either on feature engineering or classifier learning. To better mitigate the appearance diversity and ambiguity issues, we explore some new perspectives of exploiting neighborhood and background contexts in this study: 1) We view multiple pedestrian images as forming an Markov Random Field (MRF) graph in order to exploit the hidden neighborhood information for better attribute recognition performance. 
The underlying graph topology is automatically inferred, with node associations weighted by pairwise similarity between pedestrian images. The similarity can be estimated as the conventional Euclidean distance or the more elaborated decision forest-based similarity with feature selection~\cite{ZhuICCV2013,ZhuCVPR2014}.
By carrying out inference on the graph, we jointly reason and estimate the attribute probability of all images in the graph.
2) We extract foreground segments of pedestrian through deep learning-based parsing and extensively evaluate the integration of foreground segments with background context for improved pedestrian representation. All experiments are systematically conducted on the largest pedestrian attribute dataset introduced by us.

\section{Methodology}
\subsection{From Pedestrian Parsing to Representation}\label{sec:parsing}

The goal of pedestrian attribute recognition is to quantify an attribute with value, $l_\mathbf{u}$, given the $d$-dimensional feature vector, denoted by $\mathbf{u} \in \mathbb{R}^d$, of a pedestrian image.
Conventionally the features are extracted from the whole pedestrian image defined by the detection bounding box~\cite{cao2008gender,layne2012towards,Layne,Zhu}, denoted as $\mathbf{u}^\mathrm{whole}$.

However, it is more intuitive to use only the features from foreground for attribute recognition.
Would background regions play any role? 
We wish to examine if discarding the background region would facilitate more accurate recognition of pedestrian attributes. 
To this end, we train a Deep Decompositional Network (DDN)~\cite{LuoParsing} to parse a pedestrian image into different body regions.
Such a deep network is an unified architecture that combines occlusion estimation, data completion, and data transformation for pedestrian segmentation and parsing, with each layer being fully connected to the next upper layer. 
We refer readers to~\cite{LuoParsing} for the network structure and training details of the DDN due to page limits. 
At test time, the DDN parses the input image into multiple pedestrian regions.
As depicted in Fig.~\ref{fig:figParsed}, we define regions such as hair, face, body, arms, and legs of the pedestrian to be the foreground and we consider the remaining regions to be the background. 
%
%
Utilizing the binary masks (Fig.~\ref{fig:figParsed}) produced by the DDN, we investigate the following combinations of features extracted from foreground $\mathbf{u}^\mathrm{fore}$, background $\mathbf{u}^\mathrm{back}$, and the whole image $\mathbf{u}^\mathrm{whole}$, 
namely $\mathbf{u}^\mathrm{whole}$ alone, $\mathbf{u}^\mathrm{fore}$ alone, foreground and background feature concatenation $(\mathbf{u}^\mathrm{fore},\mathbf{u}^\mathrm{back})$, and foreground and whole image feature concatenation $(\mathbf{u}^\mathrm{fore},\mathbf{u}^\mathrm{whole})$.

\subsection{Recognition of Attributes using Neighborhood Context}\label{sec:mrf}

%

To improve attribute recognition, we further propose to exploit the context of neighboring images by Markov Random Field (MRF), which is an undirected graph, where each node represents a random variable and each edge represents the relation between two connected nodes. Traditionally, the neighborhood information in MRF is defined by using the nearby pixels in a single image, such as in the application of smoothing~\cite{Smoothing} in image segmentation~\cite{CoSegmentation}. In the context of attribute recognition, we hypothesize that neighboring images share natural invariance in their feature space, which could be treated as a form of regularization. As such, attribute inference of an image can be locally constrained by its neighbors to obtain a more reliable prediction. 
Hence in this work, we define the energy function of MRF over a graph $G$ as follows
\begin{equation}\label{eq:mrf}
E_{MRF}(G)=\sum_{\boldsymbol{u}\in G}C_{\boldsymbol{u}}(l_{\boldsymbol{u}})+\sum_{\boldsymbol{u}\in G}\sum_{\boldsymbol{v}\in N(\boldsymbol{u})}S_{\boldsymbol{uv}}(l_{\boldsymbol{u}},l_{\boldsymbol{v}}),
\end{equation}
where $\boldsymbol{u},\boldsymbol{v}\in G$ are two random variables in the graph and $l_{\boldsymbol{u}}$ denotes the state of $\boldsymbol{u}$. $C_{\boldsymbol{u}}$ and $S_{\boldsymbol{uv}}$ signify the unary cost and pairwise cost functions, respectively. 
More precisely, they indicate the cost of assigning state $l_{\boldsymbol{u}}$ to variable $\boldsymbol{u}$ as well as the cost of assigning states to neighboring nodes $\boldsymbol{u,v}$, which is determined based on the graph structure (e.g., assigning different states to nodes that are similar is penalized). 
$N(\boldsymbol{u})$ is a set of variables that are the neighbors of $\boldsymbol{u}$.

Each random variable corresponds to an image and the relation between two variables corresponds to the similarity between images. The variable states $l_{\boldsymbol{u}}$ are the values of the image attribute. The unary function is modeled by
\begin{equation}\label{eq:unary}
C_{\boldsymbol{u}}(l_{\boldsymbol{u}})=-\log P(l_{\boldsymbol{u}}|\boldsymbol{u}),
\end{equation}
where $P(l_{\boldsymbol{u}}|\boldsymbol{u})$ is the probability of predicting the attribute value of image $\boldsymbol{u}$ as $l_{\boldsymbol{u}}$. This probability can be conveniently mapped by the output scores of ikSVM.

Now we consider the definition of the pairwise function. To define affinity between nodes, a simple way widely adopted by existing methods, such as \cite{Zha}, is the Gaussian kernel, $\exp\{-\frac{\|\boldsymbol{u}-\boldsymbol{v}\|^2}{\sigma^2}\}$,
in which $\boldsymbol{u},\boldsymbol{v}$ indicate the feature vectors of two images and $\sigma$ is a coefficient that needs to be tuned. The graph built on this kernel function can model the global smoothness among images. However, when large variations are presented, one may consider modeling the local smoothness and discovering the intrinsic manifold of the data. Thus, an alternative is to employ the random forest (RF)~\cite{Breiman} to learn the pairwise function~\cite{ZhuICCV2013,ZhuCVPR2014}.
The RF we adopted is unsupervised, 
with pairwise sample similarity derived from the data partitioning discovered at the leaf nodes of RF as output.
The unsupervised RF can be learned using the pseudo two-class method as in~\cite{ZhuICCV2013}, \cite{ZhuCVPR2014} and \cite{liu}. 
The pairwise function in our MRF model can hence be expressed as
\begin{equation}
	S_{\boldsymbol{uv}}(l_{\boldsymbol{u}},l_{\boldsymbol{v}})) = 
    \left\{
	\begin{array}{l l}
    	\frac{1}{T}\sum_{t=1}^T\exp\{-dist^t(\boldsymbol{u},\boldsymbol{v})\} 	& \quad \text{if $l_{\boldsymbol{u}}\neq l_{\boldsymbol{v}}$,} \\
		0 	& \quad \text{otherwise.} \\
	\end{array} \right.
\label{eqn:pairwise}
\end{equation}
%
%
Here, $dist^t(\boldsymbol{u},\boldsymbol{v})=0$ if $\boldsymbol{u},\boldsymbol{v}$ fall into the same leaf node and $dist^t(\boldsymbol{u},\boldsymbol{v})$ $=+\infty$ otherwise, where $t$ is the index of tree. Since the graph is dense, the inference of MRF is difficult. Thus, we build a $k$-NN sparse graph by limiting the number of neighbors for each node. We set $k=5$ in our experiment.
Eq.(\ref{eq:mrf}) can be efficiently solved by the min-cut/max-flow algorithm introduced in \cite{min}.

\section{Experiments}
\vspace{-0.25cm}
\subsection{Settings}
\noindent \textbf{Feature representation}: Low-level color and texture features have been proven robust in describing pedestrian images~\cite{Layne}, including 8 color channels such as RGB, HSV, and YCbCr, and 21 texture channels obtained by the Gabor and Schmid filters on the luminance channel. The setting of the parameters of the Gabor and Schmid filters are given in \cite{Layne}. We horizontally partitioned the image region into six strips and then extracted the above feature channels, each of which is described by a bin-size of 16. To obtain $\mathbf{u}^\mathrm{fore}$ and  $\mathbf{u}^\mathrm{back}$, we apply the binary mask (Fig.\ref{fig:figParsed}) to extract features separately from the foreground and background.

\begin{figure}[t]
\centering
\includegraphics[width=0.5\textwidth]{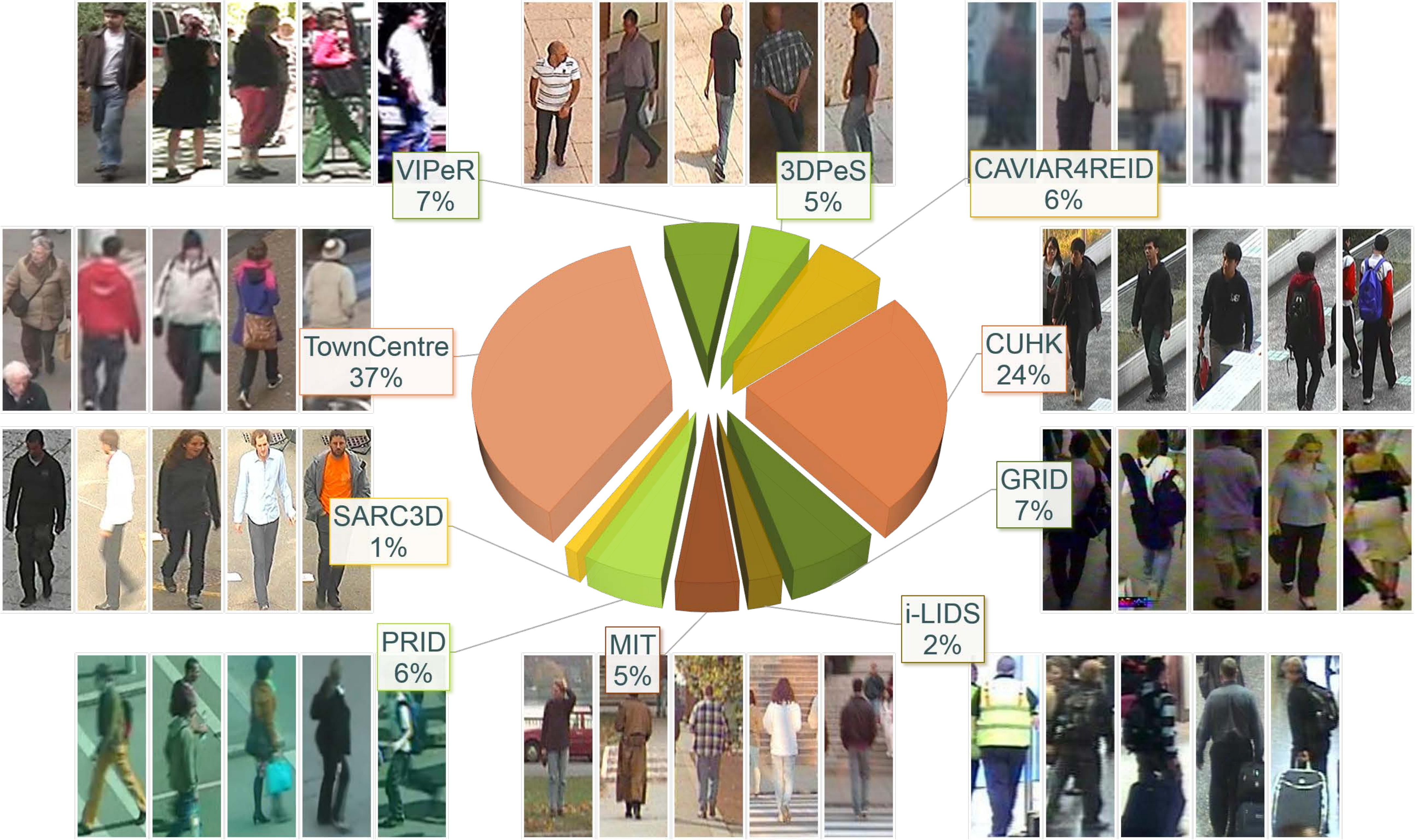}
\vskip -0.2cm
\caption{The composition of the PETA dataset.}
\vskip -0.5cm
\label{fig:pieComposition}
\end{figure}

\noindent \textbf{Dataset}: 
We present benchmark results on the PEdesTrian Attribute (PETA) dataset (Fig.~\ref{fig:pieComposition})\footnote{Dataset download: \url{http://mmlab.ie.cuhk.edu.hk/projects/PETA.html}}$^,$\footnote{Images in PETA dataset are all exclusive from those in APiS~\cite{Zhu}.} introduced by us. This dataset is the largest and most diverse pedestrian attribute dataset to date. 
There are 61 binary attributes covering an exhaustive set of characteristics of interest, including demographics (\eg~gender and age range), appearance (\eg~hair style), upper and lower body clothing style~(\eg~casual or formal), and accessories. There are another four multi-class attributes that encompass 11 basic color namings \cite{ColorName}, respectively for footwear, hair, upper-body clothing, and lower-body clothing.
We selected 35 attributes for our study, consisting of the 15 most important attributes in video surveillance proposed by human experts~\cite{Layne,HomeOfficeUK} and 20 difficult yet interesting attributes chosen by us, covering all body parts of the pedestrian and different prevalence of the attributes. For example, the attributes `sunglasses' and `v-neck' have a limited number of positive examples (Table~\ref{tb:table1}).
We randomly partitioned the dataset images into 9,500 for training, 1,900 for verification and 7,600 for testing.

\noindent \textbf{Comparisons}: 
We compare the performance of intersection kernel SVM (ikSVM)~\cite{Layne}, MRF with Gaussian kernel (MRFg), and MRF with random forest (MRFr), as discussed in Sec.\ref{sec:mrf}.
%
%
For the attributes with unbalanced positives and negatives samples, we trained ikSVM for each attribute by augmenting the positive training examples to the same size as negative examples with small variations in scale and orientation. This is to avoid bias due to imbalanced data distribution. For MRFg and MRFr, we built the graphs using two different schemes. The first scheme, symbolized by MRFg1 and MRFr1, is to construct the graphs with only the testing images. The second one, symbolized by MRFg2 and MRFr2, is to include both training and testing samples in the graphs.

\vskip -0.5cm
\subsection{Results}
\label{sec:results}

\begin{table}[t]
\centering
\caption{\footnotesize{Comparison of recognition accuracy between different feature extraction schemes. ikSVM is used as the classifier.}}
\fontsize{8}{11}\selectfont
\centering
\vskip -0.3cm
\includegraphics[width=1.88\textwidth]{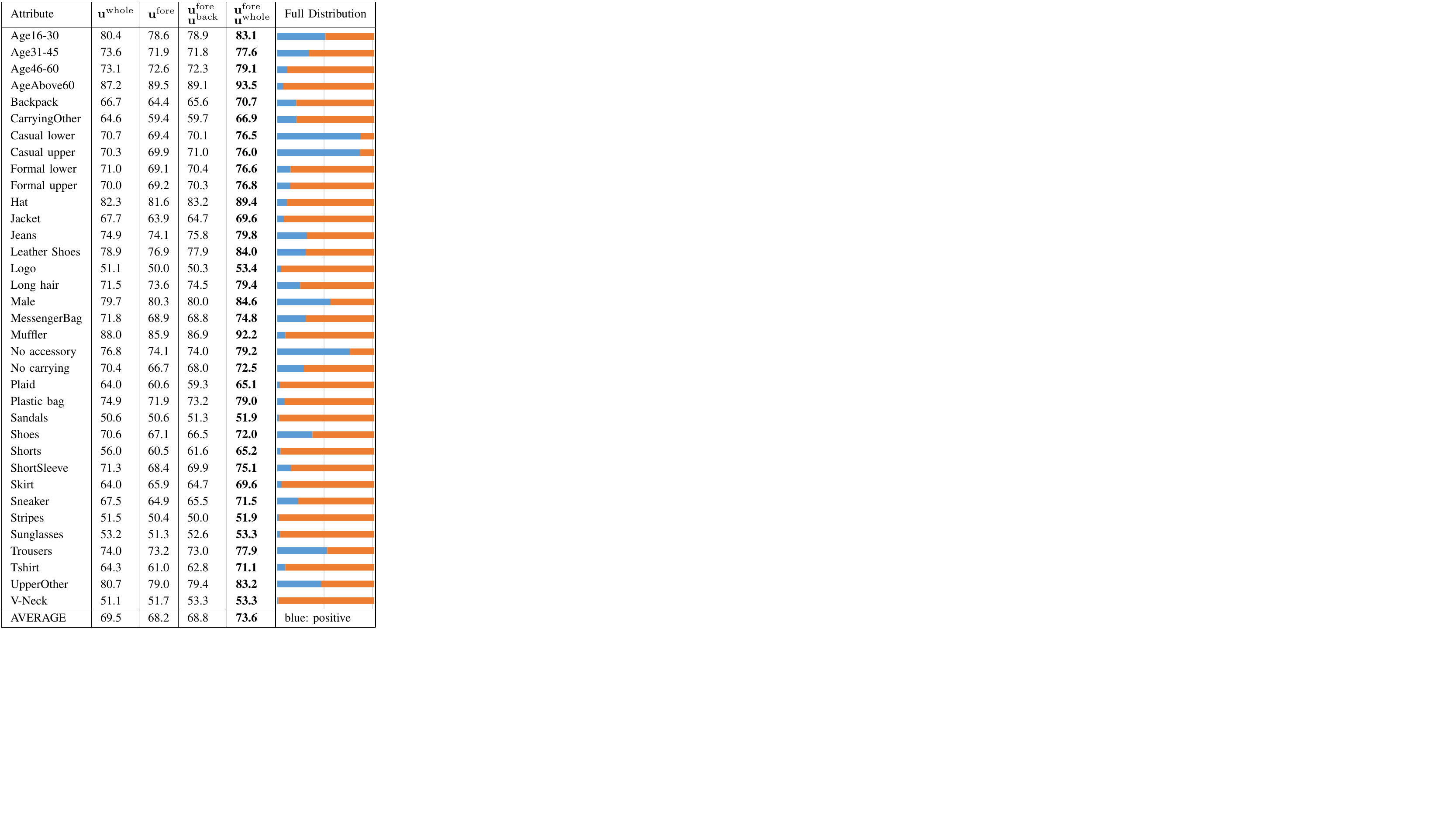}
\vskip -5.6cm
\begin{tabular}{|l|l|l|l|l|l|}
\end{tabular}
\label{tb:table1}
\end{table}

\noindent \textbf{Evaluating the informativeness of the parsed regions}:
To investigate the usefulness of foreground and background regions for attribute recognition, we first follow the previous study~\cite{layne2012towards,Layne} that applies intersection kernel SVM (ikSVM)~\cite{Maji}. 
%
%
Given the extracted foreground and background regions by DDN, we evaluate different representation schemes as discussed in Sec.~\ref{sec:parsing}, \ie~$\mathbf{u}^\mathrm{whole}$ alone, $\mathbf{u}^\mathrm{fore}$ alone, foreground and background feature concatenation $(\mathbf{u}^\mathrm{fore},\mathbf{u}^\mathrm{back})$, and foreground and whole image feature concatenation $(\mathbf{u}^\mathrm{fore},\mathbf{u}^\mathrm{whole})$.


As shown in Table~\ref{tb:table1}, we observe that simply extracting the foreground features ($\mathbf{u}^\mathrm{fore}$) results in an inferior performance than that resulted from using the whole image. It suggests that background information is critical in facilitating the detection of attributes. If we inspect the recognition results of each attribute in detail, we observe that background plays a pivotal role for recognizing `Backpack', `CarryingOther', `Plastic bag', and `No carrying' attributes. This is reasonable since the visual evidence that corresponds to these attributes is not solely captured by the pedestrian foreground region. Moreover, slight drops of accuracy are observed on cloth-style related attributes, \eg~`Jeans' and `Trousers', if features are only extracted from the foreground. These results all suggest that background region could provide context for better attribute recognition performance.

Extracting and concatenating features from the foreground and background ($(\mathbf{u}^\mathrm{fore},\mathbf{u}^\mathrm{back})$) sees a slight improvement for easy-to-spot attributes such as `AgeAbove60', `Casual upper wear', `Formal upper wear', `Hat', `Jeans', `Long hair', `Male', `Shorts', `Skirt'; however, the performance deteriorates for other attributes due to the inevitable noise contained in the features extracted from the background. 
Finally, when $(\mathbf{u}^\mathrm{fore},\mathbf{u}^\mathrm{whole})$ is adopted, a significant boost in the performance is observed, even for hard-to-spot attributes like `Leather Shoes' and `Plastic bag'.
The $(\mathbf{u}^\mathrm{fore},\mathbf{u}^\mathrm{whole})$ scheme seems to a better way to exploit the information provided by the background.

\begin{figure}[t]
\centering
\hbox{\hspace{-0ex}\includegraphics[width=.49\textwidth]{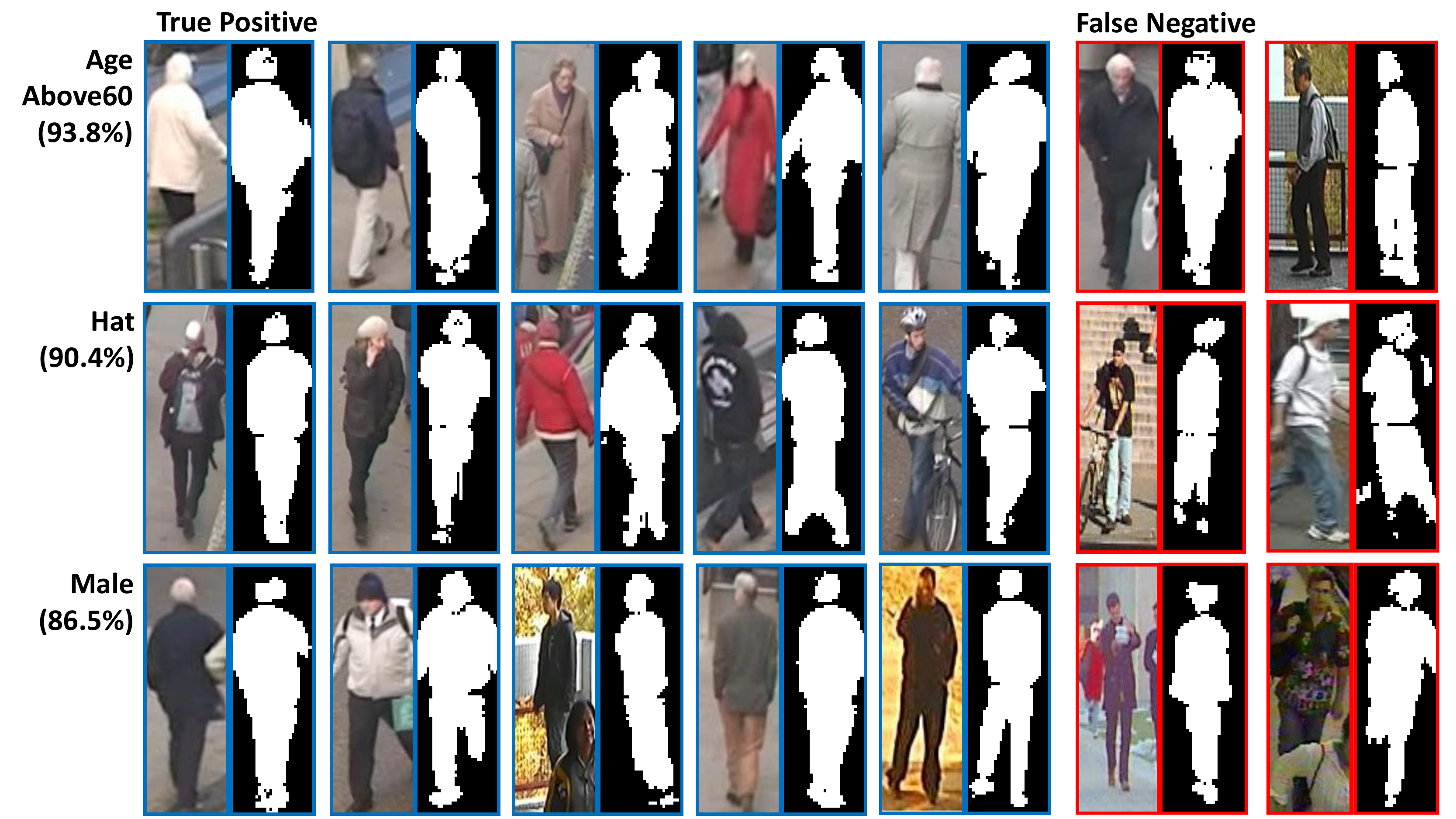}}
\hbox{\hspace{-0ex}\includegraphics[width=.49\textwidth]{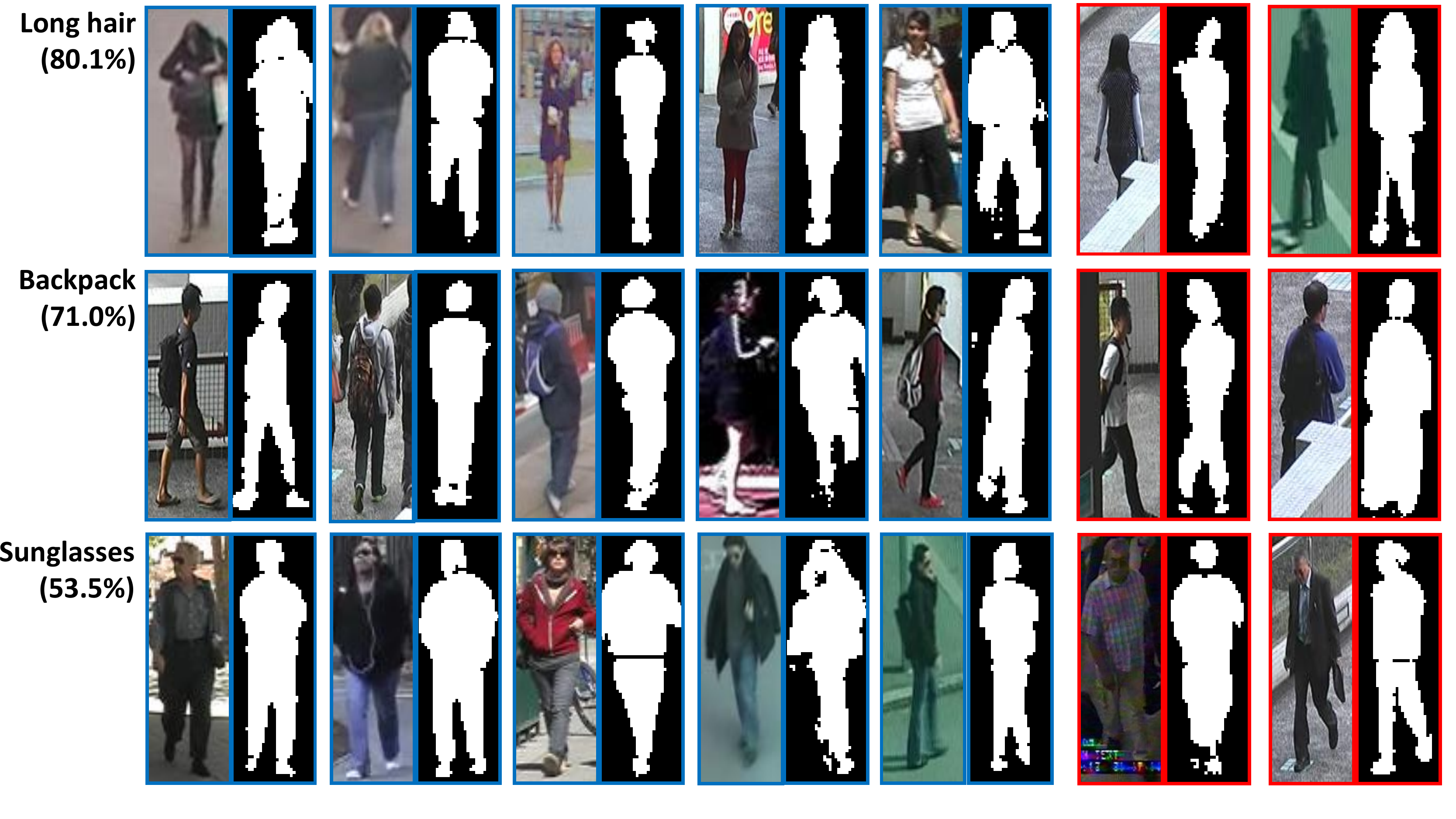}}
\vskip -0.5cm
\caption{Examples of attribute recognition with forest-based MRF (MRFr2).}
\vskip -0.6cm
\label{fig:plot}
\end{figure}

\begin{table}[h]
\centering
\caption{\footnotesize{Recognition accuracy using Markov Random Field approaches.}}
\vskip -0.3cm
\fontsize{5.8}{11}\selectfont
\begin{tabular}{|l|l|l|l|l|l|l|l|l|l|l|l|l|}

\hline
Attribute & \multicolumn{3}{l|}{MRFg1} & \multicolumn{3}{l|}{MRFg2} & \multicolumn{3}{l|}{MRFr1} & \multicolumn{3}{l|}{MRFr2} \\ \hline
Age16-30      & \multicolumn{3}{l|}{80.9          , 78.9      ,  83.2 } & \multicolumn{3}{l|}{81.7 , 78.9 , 83.2 } & \multicolumn{3}{l|}{80.9 , 81.3 , 84.8 } & \multicolumn{3}{l|}{83.8 , 83.1 , \textbf{86.8}  } \\
Age31-45      &\multicolumn{3}{l|} {74.6       , 72.3       , 78.0}       & \multicolumn{3}{l|}{76.2   , 72.3   , 78.0   }& \multicolumn{3}{l|}{74.0   , 72.2   , 80.4   }&\multicolumn{3}{l|}{ 78.8   , 76.4   , \textbf{83.1}   } \\
Age46-60      & \multicolumn{3}{l|}{74.1       , 72.6       , 79.3}       & \multicolumn{3}{l|}{75.2   , 72.6   , 79.3   }& \multicolumn{3}{l|}{73.2   , 72.6   , 78.8   }&\multicolumn{3}{l|}{ 76.4   , 75.5   , \textbf{80.1}   }\\
AgeAbove60    & \multicolumn{3}{l|}{87.2       , 89.2       , 93.4}       & \multicolumn{3}{l|}{ 88.2   , 89.2   , 93.4   }& \multicolumn{3}{l|}{86.3 ,   88.7   , 90.6   }&\multicolumn{3}{l|}{89.0   , 88.9   , \textbf{93.8}   }\\
Backpack      & \multicolumn{3}{l|}{67.1       , 65.9       , 71.0}       & \multicolumn{3}{l|}{ 67.1   , 65.9   , 71.0   }& \multicolumn{3}{l|}{67.0   , 66.0   , 70.7   }&\multicolumn{3}{l|}{ 67.2   , 66.1   , \textbf{70.5}   }\\
CarryingOther & \multicolumn{3}{l|}{64.9       , 60.2       , 67.3 }       & \multicolumn{3}{l|}{ 66.8   , 60.2   , 67.3    }& \multicolumn{3}{l|}{64.6   , 60.3   , 67.3   }&\multicolumn{3}{l|}{ 68.0   , 67.0   , \textbf{73.0}   }\\
Casual lower  & \multicolumn{3}{l|}{70.9       , 69.8       , 76.0 }       & \multicolumn{3}{l|}{ 71.6   , 69.8   , 76.1   }& \multicolumn{3}{l|}{70.4   , 69.9   , 76.2   }&\multicolumn{3}{l|}{ 71.3   , 70.9   , \textbf{78.2}   }\\
Casual upper  & \multicolumn{3}{l|}{70.4       , 70.7       , 75.4}       & \multicolumn{3}{l|}{ 71.3   , 70.7   , 75.4    }& \multicolumn{3}{l|}{69.8   , 70.3   , 75.9   }&\multicolumn{3}{l|}{ 71.3   , 71.5   , \textbf{78.1}   }\\
Formal lower  & \multicolumn{3}{l|}{71.2       , 70.9       , 76.9 }      & \multicolumn{3}{l|}{ 71.8   , 70.9   , 77.0   }& \multicolumn{3}{l|}{71.2   , 70.5   , 76.9   }&\multicolumn{3}{l|}{ 71.9   , 70.5   , \textbf{79.0}   }\\
Formal upper  & \multicolumn{3}{l|}{70.3       , 70.8       , 77.0  }      & \multicolumn{3}{l|}{ 70.4   , 70.8   , 77.1    }& \multicolumn{3}{l|}{70.3   , 70.8   , 77.1   }&\multicolumn{3}{l|}{ 70.0   , 72.0   , \textbf{78.7}   }\\
Hat           & \multicolumn{3}{l|}{82.9       , 83.2       , 89.5  }      & \multicolumn{3}{l|}{ 84.3   , 83.2   , 89.5    }& \multicolumn{3}{l|}{82.3   , 82.4   , 88.8   }&\multicolumn{3}{l|}{ 86.7   , 84.5   , \textbf{90.4}   }\\
Jacket        & \multicolumn{3}{l|}{68.3       , 65.0       , 69.8  }      & \multicolumn{3}{l|}{ 68.4   , 65.0   , 69.8    }& \multicolumn{3}{l|}{68.1   , 65.0   , 69.8   }&\multicolumn{3}{l|}{ 67.9   , 66.8   , \textbf{72.2}   }\\
Jeans         & \multicolumn{3}{l|}{75.2       , 76.3       , 80.2 }      & \multicolumn{3}{l|}{ 76.1   , 76.3   , 80.2   }& \multicolumn{3}{l|}{75.0   , 76.0   , 79.8   }&\multicolumn{3}{l|}{ 76.0   , 75.9   , \textbf{81.0}   }\\
Leather Shoes & \multicolumn{3}{l|}{80.1       , 78.0       , 84.4 }      & \multicolumn{3}{l|}{ 80.9   , 78.0   , 84.4    }& \multicolumn{3}{l|}{79.1   , 78.4   , 84.5   }&\multicolumn{3}{l|}{ 81.7   , 82.5   ,\textbf{ 87.2}   }\\
Logo          & \multicolumn{3}{l|}{51.1       , 50.5       , \textbf{53.8} }      & \multicolumn{3}{l|}{ 51.1   , 50.5   , \textbf{53.8}   }& \multicolumn{3}{l|}{51.1   , 50.0   , \textbf{53.8}   }&\multicolumn{3}{l|}{ 50.7   , 50.5   , 52.7   }\\
Long hair     & \multicolumn{3}{l|}{71.7       , 75.2       , 79.5 }      & \multicolumn{3}{l|}{ 72.6   , 75.2   , 79.5   }&\multicolumn{3}{l|}{ 71.8   , 75.1   , 79.6   }&\multicolumn{3}{l|}{ 72.8   , 75.2   , \textbf{80.1}   }\\
Male          & \multicolumn{3}{l|}{80.3       , 79.9       , 84.5 }      & \multicolumn{3}{l|}{ 80.9   , 79.9   , 84.5   }& \multicolumn{3}{l|}{80.6   , 81.3   , 85.9   }&\multicolumn{3}{l|}{ 81.4   , 81.9   , \textbf{86.5}   }\\
MessengerBag  & \multicolumn{3}{l|}{72.9       , 69.0       , 75.1 }      & \multicolumn{3}{l|}{ 74.3   , 69.0   , 75.1   }& \multicolumn{3}{l|}{72.7   , 69.0   , 74.6   }&\multicolumn{3}{l|}{ 75.5   , 73.8   , \textbf{78.3}   }\\
Muffler       & \multicolumn{3}{l|}{88.3       , 86.9       , 92.2 }      & \multicolumn{3}{l|}{ 89.5   , 86.9   , 92.2   }& \multicolumn{3}{l|}{86.5   , 86.6   , 92.3  }&\multicolumn{3}{l|}{ 91.3   , 87.9   , \textbf{93.7}   }\\
No accessory  & \multicolumn{3}{l|}{77.2       , 73.8       , 78.9 }      & \multicolumn{3}{l|}{ 78.6   , 73.8   , 78.9    }& \multicolumn{3}{l|}{77.1   , 74.8   , 79.6   }&\multicolumn{3}{l|}{ 80.0   , 78.5   ,\textbf{ 82.7}   }\\
No carrying   & \multicolumn{3}{l|}{70.6       , 68.5       , 73.1  }      & \multicolumn{3}{l|}{ 71.6   , 68.5   , 73.1  }& \multicolumn{3}{l|}{70.6   , 68.5   , 73.1  }&\multicolumn{3}{l|}{ 71.5   , 69.6   , \textbf{76.5}   }\\
Plaid         & \multicolumn{3}{l|}{64.5       , 59.6       , 65.1 }      & \multicolumn{3}{l|}{ 64.5   , 59.6   , 65.1   }& \multicolumn{3}{l|}{65.0   , 59.6   , 65.1   }&\multicolumn{3}{l|}{ 65.0   , 59.6   , \textbf{65.2}   }\\
Plastic bag   & \multicolumn{3}{l|}{74.9       , 73.6       , 79.0 }      & \multicolumn{3}{l|}{ 75.5   , 73.6   , 79.0   }& \multicolumn{3}{l|}{73.9   , 73.6   , 79.2  }&\multicolumn{3}{l|}{ 75.5   , 74.1   , \textbf{81.3}   }\\
Sandals       & \multicolumn{3}{l|}{50.6       , 51.2       , 51.6 }      & \multicolumn{3}{l|}{ 50.6   , 51.2   , 51.6   }& \multicolumn{3}{l|}{50.6   , 51.2   , 51.9   }&\multicolumn{3}{l|}{50.6   , 51.3   , \textbf{52.2}   }\\
Shoes         & \multicolumn{3}{l|}{71.0       , 66.9       , 72.4 }      & \multicolumn{3}{l|}{ 72.5   , 66.9   , 72.4   }& \multicolumn{3}{l|}{70.8   , 66.9   , 72.8   }&\multicolumn{3}{l|}{ 73.6   , 73.1   , \textbf{78.4}   }\\
Shorts        & \multicolumn{3}{l|}{56.5       , 61.8       , \textbf{65.7} }      & \multicolumn{3}{l|}{56.5   , 61.8   , \textbf{65.7}   }& \multicolumn{3}{l|}{56.5   , 61.2   , \textbf{65.7}  }&\multicolumn{3}{l|}{ 56.5   , 61.8   , 65.2   }\\
ShortSleeve   & \multicolumn{3}{l|}{71.7       , 70.5       , 75.4 }      & \multicolumn{3}{l|}{ 71.8   , 70.5   , 75.4   }& \multicolumn{3}{l|}{71.8   , 70.6   , 74.0   }&\multicolumn{3}{l|}{ 71.6   , 70.5   , \textbf{75.8}   }\\
Skirt         & \multicolumn{3}{l|}{64.0       , 65.3       , 69.6  }      & \multicolumn{3}{l|}{ 64.0   , 65.3   , 69.6   }& \multicolumn{3}{l|}{64.0   , 65.0   , \textbf{69.6}  }&\multicolumn{3}{l|}{64.3   , 65.2   , \textbf{69.6}   }\\
Sneaker       & \multicolumn{3}{l|}{68.1       , 66.2       , 72.0 }     & \multicolumn{3}{l|}{ 69.0   , 66.2   , 72.0   }& \multicolumn{3}{l|}{68.2   , 66.2   , 71.7  }&\multicolumn{3}{l|}{69.3   , 66.4   , \textbf{75.0}   }\\
Stripes       & \multicolumn{3}{l|}{\textbf{52.3}       , 50.0       ,  51.9 }     & \multicolumn{3}{l|}{ \textbf{52.3}   , 50.0   , 51.9   }& \multicolumn{3}{l|}{\textbf{52.3}   , 50.0   , 51.9   }&\multicolumn{3}{l|}{ \textbf{52.3}   , 50.0   , 51.9   }\\
Sunglasses    & \multicolumn{3}{l|}{53.2       , 52.6       ,  53.3 }      & \multicolumn{3}{l|}{ 53.2   , 52.6   , 53.3   }& \multicolumn{3}{l|}{\textbf{53.9}, 52.6   , 53.5}&\multicolumn{3}{l|}{\textbf{53.9} , 52.6   ,  53.5   }\\
Trousers      & \multicolumn{3}{l|}{74.5       , 72.9       , 77.9 }    & \multicolumn{3}{l|}{ 75.7   , 72.9   , 77.9   }& \multicolumn{3}{l|}{75.7   , 76.5   , 80.9  }&\multicolumn{3}{l|}{ 76.5   , 77.0   ,  \textbf{82.2}   }\\
Tshirt        & \multicolumn{3}{l|}{64.5       , 63.6       ,  \textbf{71.5} }     & \multicolumn{3}{l|}{ 64.6   , 63.6   ,  \textbf{71.5}  }& \multicolumn{3}{l|}{63.6   , 63.6   ,  \textbf{71.5}  }&\multicolumn{3}{l|}{ 64.2   , 63.6   , 71.4   }\\
UpperOther    & \multicolumn{3}{l|}{80.7       , 79.3       , 83.2 }      & \multicolumn{3}{l|}{ 81.8   , 79.3   , 83.2   }& \multicolumn{3}{l|}{81.1   , 81.4   , 84.3  }&\multicolumn{3}{l|}{ 83.9   , 83.3   , \textbf{ 87.3}   }\\
V-Neck        & \multicolumn{3}{l|}{51.1       , \textbf{53.3}       , \textbf{53.3} }     & \multicolumn{3}{l|}{51.1   , \textbf{53.3}   , \textbf{53.3}   }& \multicolumn{3}{l|}{51.1   , \textbf{53.3}   , \textbf{53.3}  }&\multicolumn{3}{l|}{ 51.1   , \textbf{53.3}   , \textbf{53.3}   }\\ \hline
AVERAGE       & \multicolumn{3}{l|}{69.9       , 69.0       ,  73.7  }    & \multicolumn{3}{l|}{ 70.6   , 69.0   , 73.7   }& \multicolumn{3}{l|}{69.7   , 69.2   , 73.9  }&\multicolumn{3}{l|}{71.2   , 70.6   , \textbf{75.6}  }  \\  \hline
\end{tabular}
\vskip 0.1cm
\raggedright {There are three small columns for each compared methods. They correspond to the three feature extraction}
\vskip -0.15cm
\raggedright {schemes, \ie~$\mathbf{u}^\mathrm{whole}$, $(\mathbf{u}^\mathrm{fore},\mathbf{u}^\mathrm{back})$, and $(\mathbf{u}^\mathrm{fore},\mathbf{u}^\mathrm{whole})$, respectively.}
\label{tb:table2}
\vskip -0.6cm
\end{table}

\vspace{0.1cm}
\noindent \textbf{Evaluating the importance of neighborhood context}:
We choose the best three of the four feature extraction schemes, namely the $\mathbf{u}^\mathrm{whole}$, $(\mathbf{u}^\mathrm{fore},\mathbf{u}^\mathrm{back})$, and $(\mathbf{u}^\mathrm{fore},\mathbf{u}^\mathrm{whole})$, and evaluate our proposed MRF methodology for detecting pedestrian attributes. We report the attribute detection accuracy in Table~\ref{tb:table2} and list some further observations as follows. 

Firstly, the MRF-based methods outperform ikSVM on most of the attributes (comparing Table~\ref{tb:table2} with Table~\ref{tb:table1}). For instance, MRFr2 achieves an average of $3.4\%$ improvement over ikSVM for the `age' attributes shown on top of the tables. This is significant in a dataset with large appearance diversity and ambiguity and it demonstrates that graph regularization can improve attribute inference. In addition, an about $5\%$ boost of performance is observed for attributes such as `MessengerBag', `No accessory', `No carrying', and `Trousers' and we observe a near $10\%$ boost over ikSVM for `carryingOther' and `Shoes'.
Secondly, the MRF graphs built with the second scheme (graph constructed by both train and test samples) is superior compared to the first scheme (graph constructed by test samples only), which is reasonable as using both the training and testing data can better cover the image space. Thirdly, for many important attributes, such as `Trousers' and `Shoes', random forest works much better than Gaussian kernel to measure the neighborhood context.

Moreover, we observed that for our proposed MRF methods, the importance of background information as context is best exploited 
when using $(\mathbf{u}^\mathrm{fore},\mathbf{u}^\mathrm{whole})$ (Table~\ref{tb:table2}). This observation corresponds with the detection performance using ikSVMs (Table~\ref{tb:table1}) and we show that the best result is obtained when we 
use MRFr2 with $(\mathbf{u}^\mathrm{fore},\mathbf{u}^\mathrm{whole})$, which on average outperforms the $\mathbf{u}^\mathrm{whole}$ scheme in our earlier preliminary result~\cite{deng2014pedestrian} by $4.4\%$. 
%
%
Fig.~\ref{fig:plot} shows some attribute recognition results using the forest MRF. The detection performance is satisfactory for most attributes.
False negative samples typically result from occlusion (\eg~backpack), color ambiguity (long hair) and background noise (male). All methods perform poorly on attributes with imbalanced positive-negative distribution (see Table~\ref{tb:table1}) such as `logo', `stripes', `v-neck' and `sunglasses', which are also hard to spot by human observers.


\section{Conclusions}


In this work, a novel approach to exploit the neighborhood information among image samples with emphasis on the foreground attribute regions has been investigated and the automatically inferred pairwise graph topology has led to better performance of attribute recognition. Using the latest large-scale pedestrian attribute dataset (PETA) as the benchmark, we showed that our new MRF model with the proposed feature representation scheme is more capable to accurately detect pedestrian attributes.

\bibliographystyle{abbrv}

\small{
\bibliography{ACMMM2014TCSVT}
\vskip -1cm
}

\end{document}